# PCA 4 DCA: The Application Of Principal Component Analysis To The Dendritic Cell Algorithm

Feng Gu, Julie Greensmith, Robert Oates and Uwe Aickelin
School of Computer Science
University of Nottingham, NG8 1BB, UK

## Abstract

As one of the newest members in the field of artificial immune systems (AIS), the Dendritic Cell Algorithm (DCA) is based on behavioural models of natural dendritic cells (DCs). Unlike other AIS, the DCA does not rely on training data, instead domain or expert knowledge is required to predetermine the mapping between input signals from a particular instance to the three categories used by the DCA. This data preprocessing phase has received the criticism of having manually over-fitted the data to the algorithm, which is undesirable. Therefore, in this paper we have attempted to ascertain if it is possible to use principal component analysis (PCA) techniques to automatically categorise input data while still generating useful and accurate classification results. The integrated system is tested with a biometrics dataset for the stress recognition of automobile drivers. The experimental results have shown the application of PCA to the DCA for the purpose of automated data preprocessing is successful.

## 1 Introduction

The Dendritic Cell Algorithm (DCA) is an emerging algorithm within the field of artificial immune systems (AIS) [3]. It is a biologically inspired population based algorithm which is derived from behavioural models of natural dendritic cells (DCs) [13]. It is also underpinned by a recent paradigm in immunology termed the danger theory [14], which states that the human immune system is activated in response to the detection of 'danger signals'. As an algorithm, the DCA performs fusion of real valued input signal data and correlates this information with potentially anomalous 'antigen' data. The resulting correlation values are then classified to form an anomaly detection style of two-class classification. For the algorithm to function, input signal data is classified into one of three user-defined categories. The semantics which define the categories are based on the types of input used by natural DCs, which are currently termed PAMP signal, danger signal and safe signal. For further details about the nature of the individual signal values, refer to Greensmith et al [6].

Unlike other AIS, the DCA does not rely on training data to define which of the input signals are potentially 'dangerous'. Instead, domain or expert knowledge is required in order to predetermine the mapping between input signals from a particular instance to the three categories used by the DCA. This quirk of the algorithm arose as the initial intention was to construct a gene regulatory network within each DC to process these input signals and to effect the production of related output signals. However to develop a real-time algorithm, such processing overhead was not deemed necessary and a simple weighted sum equation is employed by each cell to perform the data fusion.

This rather arbitrary and subjective method of assignment has severe limitations for the ease of application of the DCA, especially for complex and noisy datasets. It also leaves one open to the criticism of having manually over-fitted the data to the algorithm, which is undesirable. Therefore, in this paper we have attempted to use a basic feature selection method to automatically categorise input data into user-defined signal categories. The aim of this paper is to ascertain if it is possible to use principal component analysis (PCA) [12] to automatically categorise input data while still generating useful and accurate classification results. This is based on the assumption that variability of attributes is equivalent to importance. To test this hypothesis we use a biometrics dataset which aims to measure driver stress levels in city and highway locations. Section 2 of this paper describes the integration of PCA with the DCA. Section 3 describes the dataset used and how we apply it for our purpose. Section 4 describes the experiments performed, with corresponding results presented in Section 5. Finally, we present our conclusions and describe potential future avenues for this combined technique.

## 2 Integrating PCA with DCA

The DCA is a population based algorithm which performs three stages of data processing, namely signal fusion, correlation and classification. To achieve these outcomes, input signal data are categorised into PAMP, danger or safe signal categories. PAMP signals have the highest relevance to the system and are viewed as 'signatures' of anomaly. Danger signals have a high degree of anomaly associated and safe signals, as the name suggests, have a high degree of normality associated with them. The origins of these categories are rooted in the biological metaphor used initially to inspire the development of the algorithm. The mapping between application data and categories has previously arisen due to domain knowledge of the particular application. However, such mappings do not apply to the same degree if the algorithm is viewed as a computational tool. Therefore we aim to replace the subjective mapping using a systematic approach of automated signal categorisation through the application of PCA.

PCA is a mathematical operation that transforms a finite number of possibly correlated vectors into a smaller number of uncorrelated vectors, termed 'principal components'. It reveals the internal structure of given data with the focus on data variance. As a result, PCA gives not only the information of overall data variance, but also its correspondence to the variability of each vector. In addition, PCA is also used for the reduction of data dimension, by accumulating the vectors that can be linearly represented by each other.

The PCA element is used to rank input data based on the variability of each attribute. A separate ranking is also generated for the signal categories, and it is performed by using a sum of the absolute values of the weights used for signal transformation by the DCA. Once ranking is performed, the highest ranked signals from the application of PCA can be mapped to the highest ranked category and so forth.

To clarify, the process from raw data to true/false positive rates is performed in five stages as following:

1. Normalisation: All data attributes are normalised within the same range, namely [0,1], with the normalisation function dependent upon the particulars of the input data used.

2. PCA ranking and categorisation: PCA is performed on selected attributes, where attributes are ranked in terms of variability. Attributes are then categorised according to the predetermined ranking of signal categories. The highest ranked attribute forms the suspect 'antigen' data which is subsequently classified by the DCA.

3. DCA application: Signals and antigens are input to the DCA, and the $K_\alpha$ anomaly metric is generated for each antigen type.

4. Prediction Assessment: The resulting $K_\alpha$ values are divided into segments and an anomaly threshold is applied. An assessment of all $K_\alpha$ values within each segment is used to determine whether the segment is anomalous or not. A range of thresholds are used to generate ROC curves [4].

5. True/False Positive Analysis: Across all segments the true or false positive rates are calculated.

In the forthcoming sections we will describe in detail how each of these stages is performed, as well as the subsequent performance of this integrated system when applied to biometrics stress data.

## 3 Dataset And Preprocessing

The Stress Recognition in Automobile Drivers dataset [10] from the PhysioBank database [5] contains a collection of multiparameter data instances from healthy volunteers, taken while they were driving on a prescribed route including city streets and highways in and around Boston, Massachusetts. The objective of the study for which these data were collected was to investigate the feasibility of automated recognition of stress on the basis of the recorded signals, which include electrocardiogram (ECG), electromyography (EMG), galvanic skin resistance (GSR) measured on the hand and foot, heart rate (HR) and respiration.

The whole dataset consists of the data of 17 drivers, and the 'driver05' subset is used in this paper. It contains the data instances collected within a period of 5055 seconds, and each data instance includes eight attributes, which are elapsed time, ECG, EMG, foot GSR, hand GSR, HR, respiration and marker. The attribute 'maker' is seen as the indication of the changes of stress level, which is used to create labels for classification. The data instances were originally collected at an interval of 60 milliseconds, but according to [11] they are transformed into one data instance per second, by averaging the values of each attribute within every second. A summary of the statistics of all attributes is listed in Table 1. As the size of the data exceeds 5000, Quantile-Quantile Plots [2] are used to check the normality of each attribute.

|  | Min | Max | Median | Mean | StDev | Normality |
|---|---|---|---|---|---|---|
| ECG | -10.00 | 4.03 | -0.01 | -0.59 | 2.36 | no |
| EMG GSR | 0.12 | 18.24 | 0.63 | 1.25 | 1.60 | no |
| Foot GSR | 2.10 | 11.55 | 5.14 | 4.83 | 1.90 | no |
| Hand HR | 0.00 | 16.70 | 5.86 | 5.83 | 2.52 | no |
| Respiration | 0.00 | 265.00 | 71.33 | 68.13 | 19.98 | no |
| Marker | 17.69 | 50.30 | 30.35 | 29.39 | 5.97 | no |
|  | 11.47 | 60.70 | 14.35 | 15.24 | 3.10 | no |

Table 1: Statistics of the attributes in raw data.

### 3.1 Data Normalisation

The DCA requires all input signals to be in the same range, this can be achieved by performing normalisation on selected attributes. All attributes are normalised, including marker, into [0,1], which is a commonly used range in biometrics. In order to reduce the probability of gaining extra advantages through normalisation, a basic normalisation method, 'Min-Max normalisation' is employed. As the main objective of this paper is to assess the effect of applying PCA to the DCA, advanced normalisation methods may influence the classification results, making it difficult to determine whether performance changes can be contributed to the addition of PCA.

### 3.2 PCA Implementation

In order to select appropriate input signals of the DCA, PCA is performed on all attributes except for elapsed time and marker. A Biplot between the first and second principal components is displayed in Figure 1. First of all, foot GSR and hand GSR have the same rotation, which implies they are linearly representative to each other. In addition, Wilcoxon test ($p < 0.05$) [2] indicates that there is no significant difference between these two attributes. Therefore, they can be accumulated into one attribute, namely 'GSR', whose value is the mean of the two original attributes. Moreover, the variability of each attribute that contributes to the scatter of the data, that is, the correspondence between the variance of each attribute and the overall data variance, can be ranked as EMG, GSR, HR, ECG and respiration. This PCA ranking is used for antigen generation and signal categorisation.

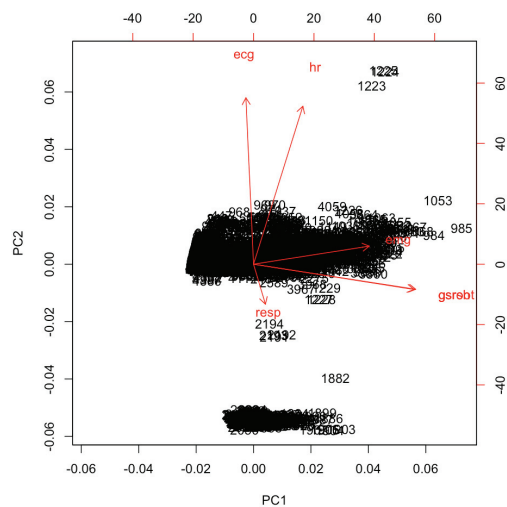

Figure 1: Biplot between the first and the second principal components.

Antigen is derived from the elapsed time of each data instance. In order to generate antigen types, a dynamic antigen multiplier [8] is employed to produce multiple identical instances of an antigen type per data instance. The amount of generated instances (antigen frequency) is determined by the attribute on the top of the PCA ranking, EMG. The antigen frequency is calculated by Equation 1.

$$F = 15 + 85 \times EMG \quad (1)$$

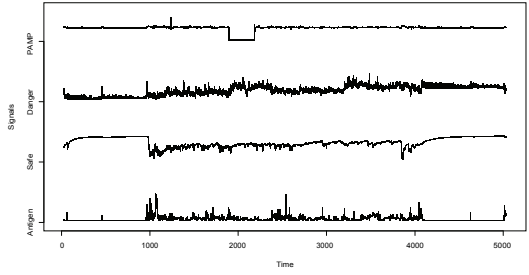

Figure 2: A plot of all signals and marker against time series, this figure does not show vertical units because each signal is scaled and offset to be shown with an illustrative amount of detail.

|  | PAMP | Danger | Safe |
|---|---|---|---|
| CSM | 2 | 1 | 2 |
| k | 2 | 1 | -3 |

Table 2: Weights for signal transformation.

This gives values between 15 and 100. The lower bound results from the fact that each data instance (per second) is accumulated from about 15 data instances (per 60 milliseconds) of the original dataset. The upper bound is based on the fact that there are 100 DCs in the population, in the most extreme case where all DCs sample the same antigen type generated from one data instance.

The rest of the attributes are mapped into the input signal categories of the DCA, by correlating the PCA ranking with the ranking of signal categories. The ranking of signal categories implies the significance of each signal category to the signal transformation of the DCA, which is in the order Safe, PAMP, and Danger. Due to the definition of Safe signals, the inverted values of the selected attribute are used. As a result, the input to the DCA is the following:

- Antigen frequency: EMG
- PAMP signal: ECG and HR
- Danger signal: respiration
- Safe signal: inverted GSR.

## 4 The Experiments

The system implemented is similar to the one demonstrated in [9], and it is programmed in C with a gcc 4.0.1 compiler. All experiments are run on an Intel 2.2 GHz MacBook (OS X 10.5.7), with the statistical tests and PCA performed in R (2.9.0). The predefined weights used for signal transformation of the DCA are displayed in Table 2, they are the same as those used in previous work [7]. The size of the DC population is set as 100, as sensitivity analyses of various population sizes [6] have shown that 100 is an appropriate value to use. The migration threshold of each DC is equal to its index multiplied by a fixed number, which produces a uniform distribution of migration thresholds. The fixed number is set to ensure the migration thresholds of most DCs in the population are greater than the strength of a single signal instance, so that these DCs can last longer than one iteration.

## 5 Results And Analysis

In order to evaluate the detection performance of the integrated system, the original dataset needs to be labelled. According to [11], the whole period being monitored can be divided into seven segments, which are in the order of 'Rest', 'City', 'Highway', 'City', 'Highway', 'City', and 'Rest'. These segments can be separated by the peaks of the attribute, marker, which was derived from human examination of video data captured at the same time as the biometric data. The segments of City are considered as those in which the driver is highly stressed, conversely the segments of Rest or Highway are considered as those in which the driver is not stressed. This can be interpreted using the terminology of anomaly detection, so the segments of City are defined as 'anomalous', whereas the segments of Rest or Highway are defined as 'normal'.

As the $K_\alpha$ values produced by the DCA share the same time series with the original dataset, the whole duration is divided into the same segments as suggested above. The plot of $K_\alpha$ values with the indication of each segment is shown in Figure 3. In order to identify whether each segment is anomalous or normal based on $K_\alpha$ values, an evaluation function is used as described in Equation 2. Assuming there is a 'segment' to be classified, $K_i$ or $K_j$ is any $K_\alpha$ value within this segment, and $Th$ is the applied threshold for classification. The true positive rate and false positive rate are calculated by comparing the classification result based on $K_\alpha$ values to the labels of the original dataset. The true pos- itive rates and false positive rates, when various thresholds applied, are listed in Table 3. To bet-ter visualise the performance of the integrated system, a ROC graph in which a ROC curve is also included as shown in Figure 4.

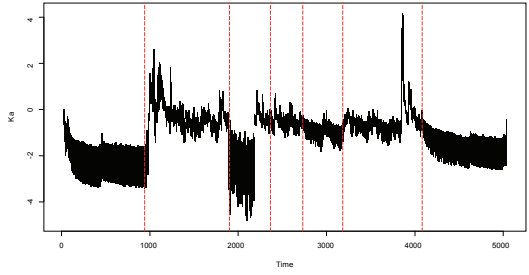

Figure 3: Plot of $K_\alpha$ values against time series that is divided into seven segments indicated by the peaks of marker.

|  | TP Rate | FP Rate |
| --- | --- | --- |
| Th = -2 | 1 | 0.75 |
| Th = -1.5 | 1 | 0.25 |
| Th = -1 | 1 | 0.25 |
| Th = -0.5 | 0.67 | 0 |
| Th = 0 | 0 | 0 |

Table 3: Results with various thresholds applied.

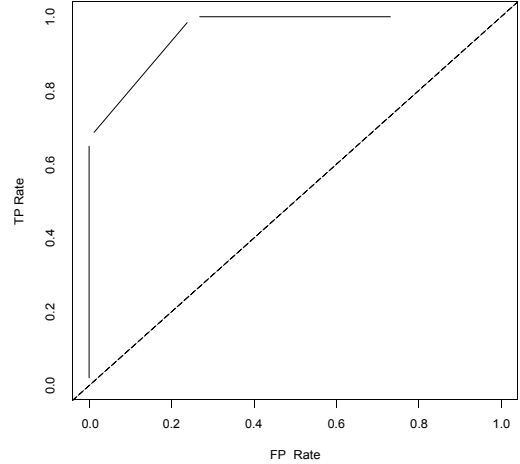

Figure 4: ROC graph of TP rates and FP rates when various thresholds are applied, the dashed line is the performance of random classifiers.

$$\text{Let } K_i \geq Th \text{ and } K_j < Th \ (i, j \in \mathbb{N})$$
$$\text{Let } L = \sum |K_i - Th| - \sum |K_j - Th| \quad (2)$$
$$\text{segment} = \begin{cases} \text{anomalous} & \text{if } L \geq 0 \\ \text{normal} & \text{otherwise} \end{cases}$$

As indicated in the ROC graph, most of the points are located in the top-left corner, which suggests the system shows high true positive rates and low false positive rates. In terms of anomaly detection, the integrated system produces good detection performance. Therefore, the application of PCA to the DCA is successful for the experimented biometrics stress data. It is possible to employ the PCA as a technique for automated data preprocessing of the DCA, without the requirement of specific expert knowledge of the problem domain.

## 6 Conclusions

We have shown that it is possible to integrate PCA with the DCA for the purpose of automated data preprocessing. The PCA facilitates the reduction of data dimension of the raw data, to select proper attributes as the candidates of the input of the DCA. It is also used for the ranking of attributes based on the variability, which is mapped to the ranking of signal categories of the DCA for signal categorisation. In this way, the data preprocessing of the DCA is performed by simply using PCA and basic Min-Max normalisation, without requiring any expert knowledge of the problem domain. The results suggest that the integrated system of PCA and the DCA is successful in terms of anomaly detection, as the system can produce relatively high true positive rates and low false positive rates. As a result, the application of PCA to the DCA makes it possible to automatically categorise input data into user-defined signal categories, while still generating useful and accurate classification results. The hypothesis is tested to be true.

Since the integrated system is automated without any human intervention during detection, it is possible to apply the system to real-time detection tasks. As the data are collected during detection, the system can use PCA to perform signal categorisation on the current batch of data, in order to generate the input data of the DCA. The DCA then performs anomaly detection on the input data, to produce the final detection results in which anomalies within the collected data can be identified. As new incoming data are collected, the system repeats the process until a termination condition is reached. Therefore, periodic detection can be performed in such an integrated system, which is essential for a real-time detection system.

The main objective of this paper is to introduce the possibility of using statistical techniques to perform automated data processing of the DCA. PCA is one of many available options, but other techniques, such as N-Gram analysis [1] and so on, yield beneficial results. Moreover, the integrated system is only tested on one particular biometrics dataset, in order to further validate the system, it needs to be applied to other biometrics datasets with similar features.